%% file: pedestrian.tex
\documentclass[10pt,twocolumn,letterpaper]{article}

\usepackage{iccv}
\usepackage{times}
\usepackage{epsfig}
\usepackage{graphicx}
\usepackage{amsmath}
\usepackage{amssymb}
\usepackage{enumitem}
\usepackage{lipsum}
\usepackage{enumerate}

\usepackage[pagebackref=true,breaklinks=true,letterpaper=true,colorlinks,bookmarks=false]{hyperref}

\newcommand{\Sim}{\mathord{\sim}}

\def\eqnvspace{{\vspace{-3mm}}}
\def\figvspace{{\vspace{-4mm}}}
\newcommand{\Paragraph}[1]{\vspace{0.5mm}\noindent\textbf{#1}\hspace{-0mm}}
\newcommand{\Section}[1]{\vspace{-2mm} \section{#1} \vspace{-1mm}}

\iccvfinalcopy 

\def\iccvPaperID{2435} 

\ificcvfinal\fi

\begin{document}

\title{Illuminating Pedestrians via Simultaneous Detection \& Segmentation}

\author{Garrick Brazil, Xi Yin, Xiaoming Liu \\
Michigan State University, East Lansing, MI 48824\\
{\tt\small \{brazilga, yinxi1, liuxm\}@msu.edu}
}

\maketitle
\thispagestyle{empty}

\begin{abstract}

Pedestrian detection is a critical problem in computer vision with significant impact on safety in urban autonomous driving. In this work, we explore how semantic segmentation can be used to boost pedestrian detection accuracy while having little to no impact on network efficiency. We propose a segmentation infusion network to enable joint supervision on semantic segmentation and pedestrian detection. When placed properly, the additional supervision helps guide features in shared layers to become more sophisticated and helpful for the downstream pedestrian detector. Using this approach, we find weakly annotated boxes to be sufficient for considerable performance gains. We provide an in-depth analysis to demonstrate how shared layers are shaped by the segmentation supervision. In doing so, we show that the resulting feature maps become more semantically meaningful and robust to shape and occlusion. Overall, our simultaneous detection and segmentation framework achieves a considerable gain over the state-of-the-art on the Caltech pedestrian dataset, competitive performance on KITTI, and executes  $2\times$ faster than competitive methods.

\end{abstract}

\input{sec_1.tex}
\input{sec_2.tex}
\input{sec_3_gb.tex}
\input{sec_4.tex}
\input{sec_5.tex}

{
\small
\bibliographystyle{ieee}
\bibliography{egbib}
}

\end{document}

%% file: sec_1.tex
\section{Introduction}

Pedestrian detection from an image is a core capability of computer vision, due to its applications such as autonomous driving and robotics~\cite{Geiger2012CVPR}. 
It is also a long-standing vision problem because of its distinct challenges including low resolution, occlusion, cloth variations, etc~\cite{zhang2016far}. 
There are two central approaches for detecting pedestrians: object detection~\cite{cai2016unified, zhang2016faster} and semantic segmentation~\cite{chen2016deeplab, cordts2016cityscapes}. 
The two approaches are highly related by nature but have their own strengths and weaknesses. 
For instance, object detection is designed to perform well at localizing distinct objects but typically provides little information on object boundaries. In contrast, semantic segmentation does well at distinguishing pixel-wise boundaries among classes but struggles to separate objects within the same class. 

\begin{figure}[t]
\begin{center}
   \includegraphics[width=1\linewidth]{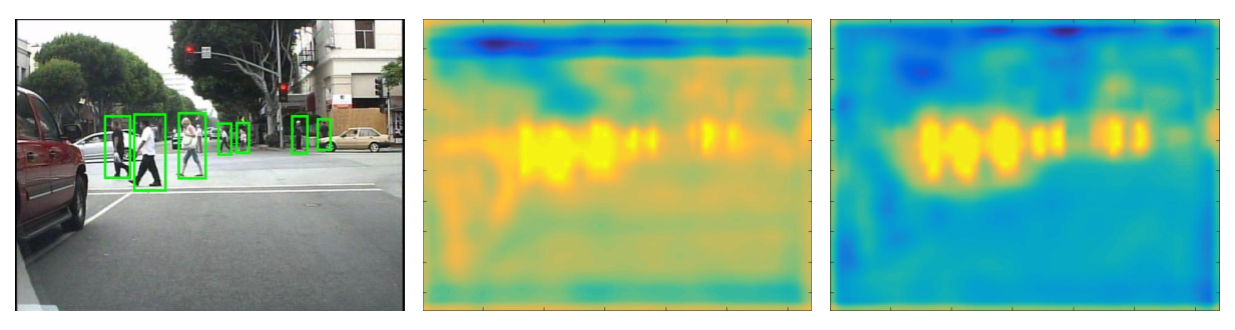}
\end{center}\eqnvspace
   \caption{Detection results on the Caltech test set (left), feature map visualization from the RPN of conventional Faster R-CNN (middle), and feature map visualization of SDS-RCNN (right). Notice that our feature map substantially illuminates the pedestrian shape while suppressing the background region, both of which make positive impact to downstream pedestrian detection.}
\label{fig:intro_fig}\figvspace
\end{figure}

Intuitively, we expect that knowledge from either task will make the other substantially easier. 
This has been demonstrated for generic object detection, since having segmentation masks of objects would clearly facilitate detection. 
For example, Fidler et al.~\cite{fidler2013bottom} utilize predicted segmentation masks to boost  object detection performance via a deformable part-based model. 
Hariharan et al.~\cite{hariharan2014simultaneous} show how segmentation masks generated from MCG~\cite{arbelaez2014multiscale} can be used to mask background regions and thus simplify detection. 
Dai et al.~\cite{dai2016instance} utilize the two tasks in a $3$-stage cascaded network consisting of box regression, foreground segmentation, and classification. 
Their architecture allows each task to share features and feed into one another. 

In contrast, the pairing of these two tasks is rarely studied in pedestrian detection, despite the recent advances~\cite{cai2016unified,li2015scale,zhang2016faster}.
This is due in part to the lack of pixel-wise annotations available in classic pedestrian datasets such as Caltech~\cite{dollar2009pedestrian} and KITTI~\cite{Geiger2012CVPR}, unlike the detailed segmentation labels in the COCO~\cite{lin2014microsoft} dataset for generic object detection.
With the release of Cityscapes~\cite{cordts2016cityscapes}, a high quality dataset for urban semantic segmentation, it is expected that substantial research efforts will be on {\it how to leverage semantic segmentation to boost the performance of pedestrian detection}, which is the core problem to be studied in this paper.

Given this objective, we start by presenting a competitive two-stage baseline framework of pedestrian detection deriving from RPN+BF~\cite{zhang2016faster} and Faster~R-CNN~\cite{ren2015faster}. 
We contribute a number of key changes to enable the second-stage classifier to specialize in stricter supervision and additionally fuse the refined scores with the first stage RPN. 
These changes alone lead to state-of-the-art performance on the Caltech benchmark. 
We further present a simple, but surprisingly powerful, scheme to utilize multi-task learning on pedestrian detection and semantic segmentation. 
Specifically, we infuse the semantic segmentation mask into shared layers using a \textit{segmentation infusion layer} in both stages of our network. 
We term our approach as ``simultaneous detection and segmentation R-CNN (SDS-RCNN)".
We provide an in-depth analysis on the effects of joint training by examining the shared feature maps, e.g., Fig.~\ref{fig:intro_fig}.
Through infusion, the shared feature maps begin to illuminate pedestrian regions.
Further, since we infuse the semantic features during training only, the network efficiency at inference is unaffected. 
We demonstrate the effectiveness of SDS-RCNN by reporting considerable improvement ($23\%$ relative reduction of the error) over the published state-of-the-art on Caltech~\cite{dollar2009pedestrian}, competitive performance on KITTI~\cite{Geiger2012CVPR}, and a runtime roughly $2\times$ faster than competitive methods.

In summary our contributions are as follows:
\begin{itemize}[noitemsep,topsep=1mm,label=$\diamond$]
\setlength\itemsep{1mm}
\item Improved baseline derived from~\cite{ren2015faster, zhang2016faster} by enforcing stricter supervision in the second-stage classification network, and further fusing scores between stages. 
\item A multi-task infusion framework for joint supervision on pedestrian detection and semantic segmentation, with the goal of \textit{illuminating} pedestrians in shared feature maps and easing downstream classification.	 
\item  We achieve the new state-of-the-art performance on Caltech pedestrian dataset, competitive performance on KITTI, and obtain $2\times$ faster runtime. 
\end{itemize}

%% file: sec_2.tex
\section{Prior work}

\Paragraph{Object Detection:} Deep convolution neural networks have had extensive success in the domain of object detection. 
Notably, derivations of Fast~\cite{girshick2015fast} and Faster~R-CNN~\cite{ren2015faster} are widely used in both generic object detection~\cite{cai2016unified, gidaris2015object, yang2016exploit} and pedestrian detection~\cite{li2015scale, tian2015deep, zhang2016faster}. 
Faster R-CNN consists of two key components: a region proposal network (RPN) and a classification sub-network. 
The RPN works as a sliding window detector by determining the \textit{objectness} across a set of predefined anchors (box shapes defined by aspect ratio and scale) at each spatial location of an image. 
After object proposals are generated, the second stage classifier determines the precise class each object belongs to. 
Faster R-CNN has been shown to reach state-of-the-art performance on the PASCAL VOC 2012~\cite{pascalvoc2012} dataset for generic object detection and continues to serve as a frequent baseline framework for a variety of related problems~\cite{gidaris2015object, hariharan2014simultaneous, he2016deep, zhang2016far}.

\Paragraph{Pedestrian Detection:} 
Pedestrian detection is one of the most extensively studied problems in object detection due to its real-world significance. 
The most notable challenges are caused by small scale, pose variations, cyclists, and occlusion~\cite{zhang2016far}. 
For instance, in the Caltech pedestrian dataset~\cite{dollar2009pedestrian} $70\%$ of pedestrians are occluded in at least one frame.

The top performing approaches on the Caltech pedestrian benchmark are variations of Fast or Faster R-CNN. 
SA-FastRCNN~\cite{girshick2015fast} and MS-CNN~\cite{cai2016unified} reach competitive performance by directly addressing the scale problem using specialized multi-scale networks integrated into Fast and Faster R-CNN respectively. 
Furthermore, RPN+BF~\cite{zhang2016faster} shows that the RPN of Faster R-CNN performs well as a standalone detector while the downstream classifier degrades performance due to collapsing bins of small-scale pedestrians. 
By using higher resolution features and replacing the downstream classifier with a boosted forest, RPN+BF is able to alleviate the problem and achieve $9.58\%$ miss rate on the Caltech reasonable~\cite{dollar2012pedestrian} setting.
F-DNN~\cite{du2016fused} also uses a derivation of the Faster R-CNN framework. 
Rather then using a single downstream classifier, F-DNN fuses multiple parallel classifiers including ResNet~\cite{he2016deep} and GoogLeNet~\cite{szegedy2015going} using soft-reject and further incorporates multiple training datasets to achieve $8.65$\% miss rate on the Caltech reasonable setting. 
The majority of top performing approaches utilize some form of a RPN, whose scores are typically discarded after selecting the proposals.
In contrast, our work shows that fusing the score with the second stage network can lead to substantial performance improvement.

\Paragraph{Simultaneous Detection \& Segmentation:} 
There are two lines of research on simultaneous detection and segmentation.
The first aims to improve the performance of both tasks, and formulates a problem commonly known as \textit{instance-aware semantic segmentation}~\cite{cordts2016cityscapes}. 
Hariharan et al.~\cite{hariharan2014simultaneous} predict segmentation masks using MCG~\cite{arbelaez2014multiscale} then get object instances using ``slow" R-CNN~\cite{girshick2014rich} on masked image proposals. 
Dai et al.~\cite{dai2016instance} achieve high performance on instance segmentation using an extension of Faster R-CNN in a $3$-stage cascaded network including mask supervision. 

\begin{figure*}[t]
\begin{center}
   \includegraphics[width=.90\linewidth]{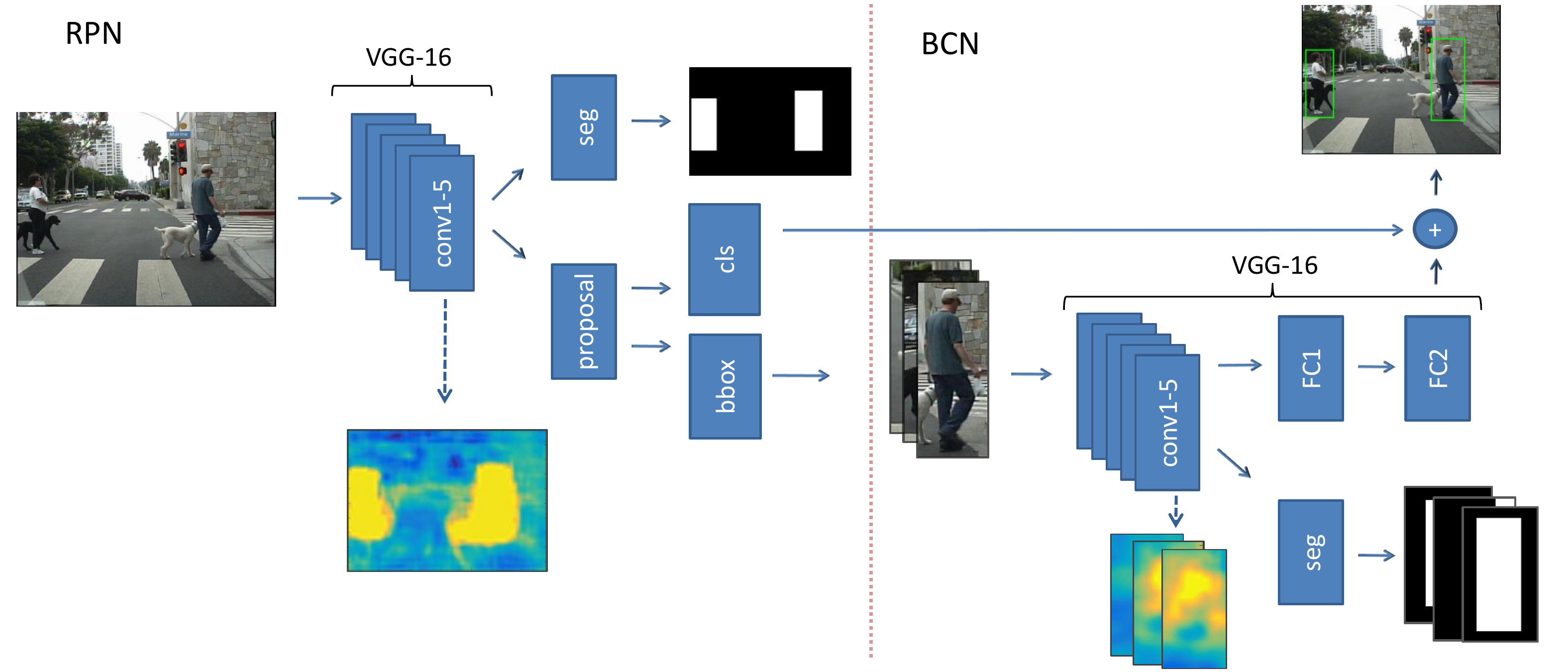}
\end{center}\eqnvspace
   \caption{Overview of the proposed SDS-RCNN framework. The segmentation layer infuses semantic features into shared conv1-5 layers of each stage, thus \textit{illuminating} pedestrians and easing downstream pedestrian detection (proposal layers in RPN, and FC1-2 in BCN).  } 
\label{fig:network}\figvspace
\end{figure*}

The second aims to explicitly improve object detection by using segmentation as a strong cue.
Early work on the topic by Fidler et al.~\cite{fidler2013bottom} demonstrates how semantic segmentation masks can be used to extract strong features for improved object detection via a deformable part-based model. 
Du et al.~\cite{du2016fused} use segmentation as a strong cue in their F-DNN+SS framework.
Given the segmentation mask predicted by a third parallel network, their ensemble network uses the mask in a post-processing manner to suppress background proposals, and pushes performance on the Caltech pedestrian dataset from $8.65\%$ to $8.18\%$ miss rate. 
However, the segmentation network degrades the efficiency of F-DNN+SS from $0.30$ to $2.48$ seconds per image, and requires multiple GPUs at inference. 
In contrast, our novel framework infuses the semantic segmentation masks into shared feature maps and thus does not require a separate segmentation network, which outperforms~\cite{du2016fused} in both accuracy and network efficiency. 
Furthermore, our use of weak box-based segmentation masks addresses the issue of lacking pixel-wise segmentation annotations in~\cite{dollar2009pedestrian, Geiger2012CVPR}.


%% file: sec_3_gb.tex
\section{Proposed method}

Our proposed architecture consists of two key stages: a region proposal network (RPN) to generate candidate bounding boxes and corresponding scores, and a binary classification network (BCN) to refine their scores. 
In both stages, we propose a \textit{semantic segmentation infusion layer} with the objective of making downstream classification a substantially easier task.
The infusion layer aims to encode semantic masks into shared feature maps which naturally serve as strong cues for pedestrian classification. 
Due to the impressive performance of the RPN as a standalone detector, we elect to fuse the scores between stages rather than discarding them as done in prior work~\cite{cai2016unified, du2016fused, tsogkas2015deep, zhang2016faster}.
An overview of the SDS-RCNN framework is depicted in Fig.~\ref{fig:network}

\subsection{Region Proposal Network}

The RPN aims to propose a set of bounding boxes with associated confidence scores around potential pedestrians. 
We adopt the RPN of Faster R-CNN~\cite{ren2015faster} following the settings in~\cite{zhang2016faster}. 
We tailor the RPN for pedestrain detection by configuring $N_a = 9$ anchors with a fixed aspect ratio of $0.41$ and spanning a scale range from $25$~--~$350$ pixels, corresponding to the pedestrain statistics of Caltech~\cite{dollar2009pedestrian}.
Since each anchor box acts as a sliding window detector across a pooled image space, there are $N_p = N_a \times \frac{W}{f_s} \times \frac{H}{f_s}$ total pedestrian proposals, where $f_s$ corresponds to the feature stride of the network.
Hence, each proposal box $i$ corresponds to an anchor and a spatial location of image $\bf{I}$.
 
The RPN architecture uses conv1-5 from VGG-16~\cite{simonyan2014very} as the backbone.
Following \cite{ren2015faster}, we attach a proposal feature extraction layer to the end of the network with two sibling output layers for box classification (\textit{cls}) and bounding box regression (\textit{bbox}). 
We further add a segmentation infusion layer to conv5 as detailed in Sec.~\ref{sec:seg}. 

For every proposal box $i$, the RPN aims to minimize the following joint loss function with three terms:
\begin{equation}
\eqnvspace
L = \lambda_c\sum_iL_{c}(c_i, \hat{c_i}) + \lambda_r\sum_iL_{r}(t_i, \hat{t_i}) + \lambda_s L_{s}.
\eqnvspace
\end{equation}
The first term is the classification loss $L_{c}$, which is a softmax logistic loss over two classes (pedestrian vs.~background).
We use the standard labeling policy which considers a proposal box at location $i$ to be pedestrian ($c_i=1$) if it has at least $0.5$ Intersection over Union (IoU) with a ground truth pedestrian box, and otherwise background ($c_i = 0$).
The second term seeks to improve localization via bounding box regression, which learns a transformation for each proposal box to the nearest pedestrian ground truth.
Specifically, we use $L_{r}(t_i, \hat{t_i})=R(t_i- \hat{t_i})$ where $R$ is the robust (smooth $L_1$) loss defined in~\cite{girshick2015fast}.
The bounding box transformation is defined as a $4$-tuple consisting of shifts in $x$, $y$ and scales in $w$, $h$ denoted as $t = [t_x, t_y, t_w, t_h]$.
The third term $L_s$ is the segmentation loss presented in Sec.~\ref{sec:seg}.

In order to reduce multiple detections of the same pedestrian, we apply non-maximum suppression (NMS) greedily to all pairs of proposals
after the transformations have been applied.
We use an IoU threshold of $0.5$ for NMS.

We train the RPN in the Caffe~\cite{jia2014caffe} framework using SGD with a learning rate of $0.001$, momentum of 0.9, and mini-batch of $1$ full-image.
During training, we randomly sample $120$ proposals per image at a ratio of $1$:$5$ for pedestrian and background proposals to help alleviate the class imbalance.
All other proposals are treated as ignore.
We initialize conv1-5 from a VGG-16 model pretrained on ImageNet~\cite{deng2009imagenet}, and all remaining layers randomly.
Our network has four max-pooling layers (within conv1-5), hence $f_s=16$.
In our experiments, we regularize our multi-task loss terms by setting $\lambda_c = \lambda_s = 1, \lambda_r = 5$.

\subsection{Binary Classification Network}
\label{sec:bcn}

\begin{figure}[t]
\begin{center}
    \includegraphics[trim={190 335 90mm 55mm},clip,  width=0.9\linewidth]{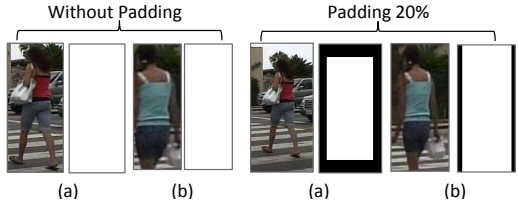}
\end{center}\eqnvspace
   \caption{Example proposal masks with and without padding. There is no discernible difference between the non-padded masks of well-localized (a) and poorly localized (b) proposals. }
\label{fig:padding}\figvspace
\end{figure}

The BCN aims to perform pedestrian classification over the proposals of the RPN. 
For generic object detection, the BCN usually uses the downstream classifier of Faster R-CNN by sharing conv1-5 with the RPN, but was shown by \cite{zhang2016faster} to degrade pedestrian detection accuracy. 
Thus, we choose to construct a separate network using VGG-16.
The primary advantage of a separate network is to allow the BCN freedom to specialize in the types of ``harder" samples left over from the RPN. 
While sharing computation is highly desirable for the sake of efficiency, the shared networks are more predestined to predict similar scores which are redundant when fused. 
Therefore, rather than cropping and warping a shared feature space, our BCN directly crops the top $N_b$ proposals from the RGB input image.

For each proposal image $i$, the BCN aims to minimize the following joint loss function with two terms:
\begin{equation}
\label{eqn:jointloss}\eqnvspace
L = \lambda_c\sum_iw_iL_{c}(c_i, \hat{c_i}) + \lambda_s L_{s}.
\eqnvspace
\end{equation}
Similar to RPN, the first term is the classification loss $L_c$ where ${c}_i$ is the class label for the $i$th proposal.
A cost-sensitive weight $w_i$ is used to give precedence to detect large pedestrians over small pedestrians. 
There are two key motivations for this weighting policy. First, large pedestrians typically imply close proximity and are thus significantly more important to detect. 
Secondly, we presume that features of large pedestrians may be more helpful for detecting small pedestrians. 
We define the weighting function given the $i$th proposal with height $h_i$ and a pre-computed mean height $\bar{h}$ as $w_i = 1 + \frac{h_i}{\bar{h}}$.
The second term is the segmentation loss presented in Sec.~\ref{sec:seg}. 

We make a number of significant contributions to the BCN. 
First, we change the labeling policy to encourage higher precision and further diversification from the RPN. 
We enforce a \textit{stricter} labeling policy, requiring a proposal to have IoU $> 0.7$ with a ground truth pedestrian box to be considered pedestrian ($c_i = 1$), and otherwise background ($c_i = 0$).
This encourages the network to suppress poorly localized proposals and reduces false positives in the form of double detections. 
Secondly, we choose to fuse the scores of the BCN with the confidence scores of the RPN at test time. 
Since our design explicitly encourages the two stages to diversify, we expect the classification characteristics of each network to be complementary when fused.
We fuse the scores at the feature level prior to softmax.
Formally, the fused score for the $i$th proposal, given the predicted $2$-class scores from the RPN $ = \{\hat{c}_{i0}^{r}$, $\hat{c}_{i1}^{r}\}$ and BCN $ = \{\hat{c}_{i0}^{b}$, $\hat{c}_{i1}^{b}\}$ is computed via the following softmax function:

\begin{equation}
\eqnvspace
\hat{c_{i}} = \frac{e^{(\hat{c}_{i1}^{r} + \hat{c}_{i1}^{b})}}{e^{(\hat{c}_{i1}^{r} + \hat{c}_{i1}^{b})} + e^{(\hat{c}_{i0}^{r} + \hat{c}_{i0}^{b})}}
\eqnvspace.
\end{equation}
In effect, the fused scores become more confident when the stages agree, and otherwise lean towards the dominant score.
Thus, it is ideal for each network to diversify in its classification capabilities such that at least one network may be \textit{very} confident for each proposal.

For a modest improvement to efficiency, we remove the pool5 layer from the VGG-16 architecture then adjust the input size to $112\times112$ to keep the fully-connected layers intact.
This is a fair trade-off since most pedestrian heights fall in the range of $30-80$~ pixels~\cite{dollar2009pedestrian}. 
Hence, small pedestrian proposals are upscaled by a factor of $\Sim2\times$, allowing space for finer discrimination.
We further propose to pad each proposal by $20\%$ on all sides to provide background context and avoid partial detections, as shown in Fig.~\ref{fig:padding}.

We train the BCN in the Caffe~\cite{jia2014caffe} framework using the same settings as the RPN.
We initialize conv1-5 from the trained RPN model, and all remaining layers randomly.
During training, we set $N_b=20$. 
During inference, we set $N_b=15$ for a moderate improvement to efficiency. 
We regularize the multi-task loss by setting $\lambda_c = \lambda_s = 1$.

\subsection{Simultaneous Detection \& Segmentation}
\label{sec:seg}

\begin{figure}[t!]
\begin{center}
   \includegraphics[width=1\linewidth]{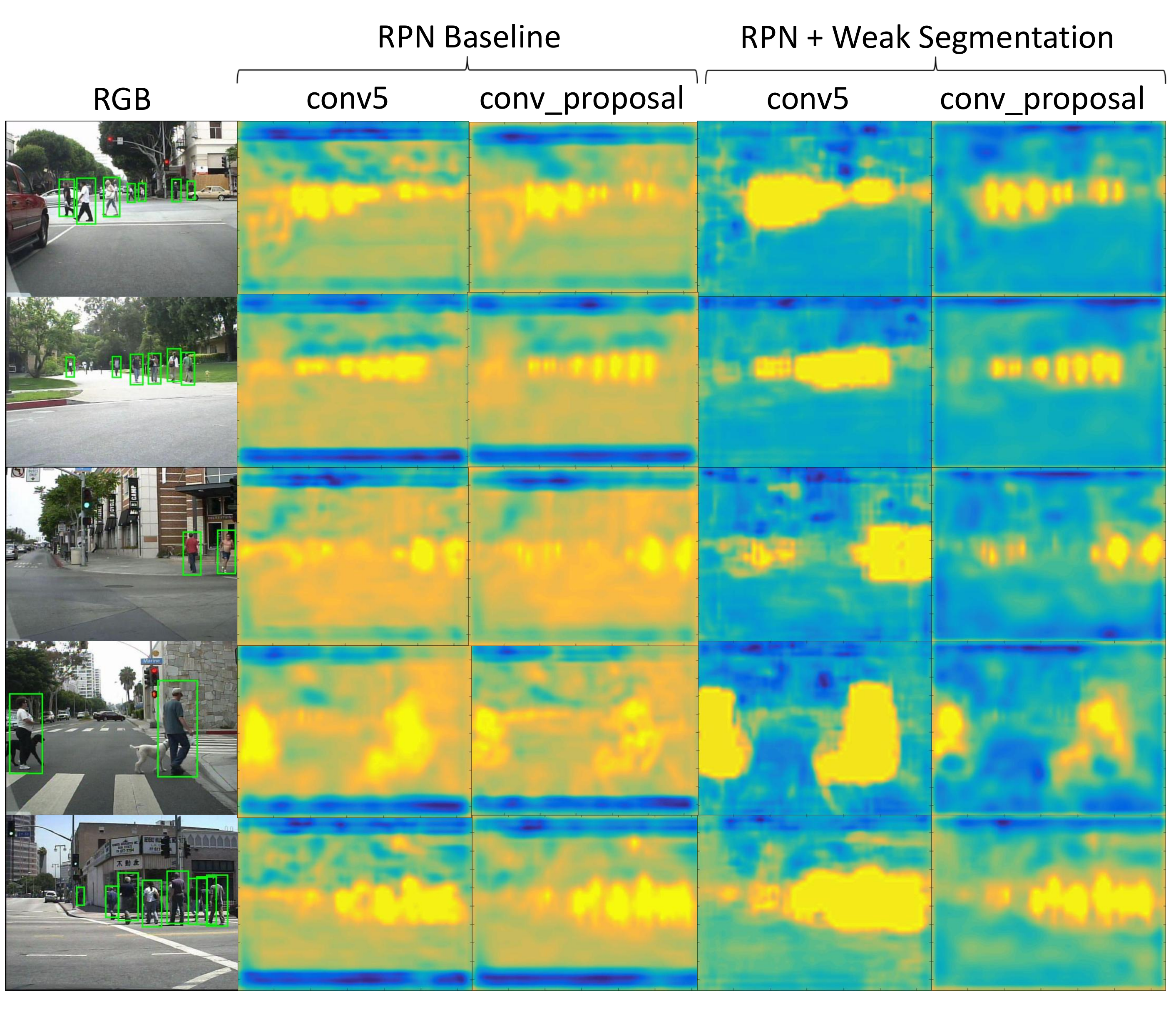}
\end{center}
\eqnvspace
   \caption{Feature map visualizations of conv5 and the proposal layer for the baseline RPN (left) and the RPN infused with weak segmentation supervision (right).
}
\label{fig:samples}\figvspace
\end{figure}

We approach simultaneous detection and segmentation with the motivation to make our downstream pedestrian detection task easier. 
We propose a segmentation infusion layer trained on weakly annotated pedestrian boxes which \textit{illuminate} pedestrians in the shared feature maps preceding the classification layers.
We integrate the infusion layer into both stages of our SDS-RCNN framework.

\Paragraph{Segmentation Infusion Layer:}
The segmentation infusion layer 
aims to output two masks indicating the likelihood of residing on pedestrian or background segments.
We choose to use only a single layer and a $1\times1$ kernel so the impact on the shared layers will be as high as possible.
This forces the network to directly infuse semantic features into shared feature maps, as visualized in Fig.~\ref{fig:samples}.
A deeper network could achieve higher segmentation accuracy but will infer less from shared layers and  diminish the overall impact on the downstream pedestrian classification.
Further, we choose to attach the infusion layer to conv5 since it is the deepest layer which precedes both the proposal layers of the RPN and the fully connected layers of the BCN. 

Formally, the final loss term $L_{s}$ of both the RPN and BCN is a softmax logistic loss over two classes (pedestrian vs.~background), applied to each location $i$, where $w_i$ is the cost-sensitive weight introduced in \ref{sec:bcn}:


\begin{equation}
\label{eqn:segloss}\eqnvspace
\lambda_s\sum_iw_iL_{s}(\mathbf{S}_i, \hat{\mathbf{S}_i}).
\eqnvspace
\end{equation}



\begin{figure}[t]
\begin{center}
   \includegraphics[width=.9\linewidth]{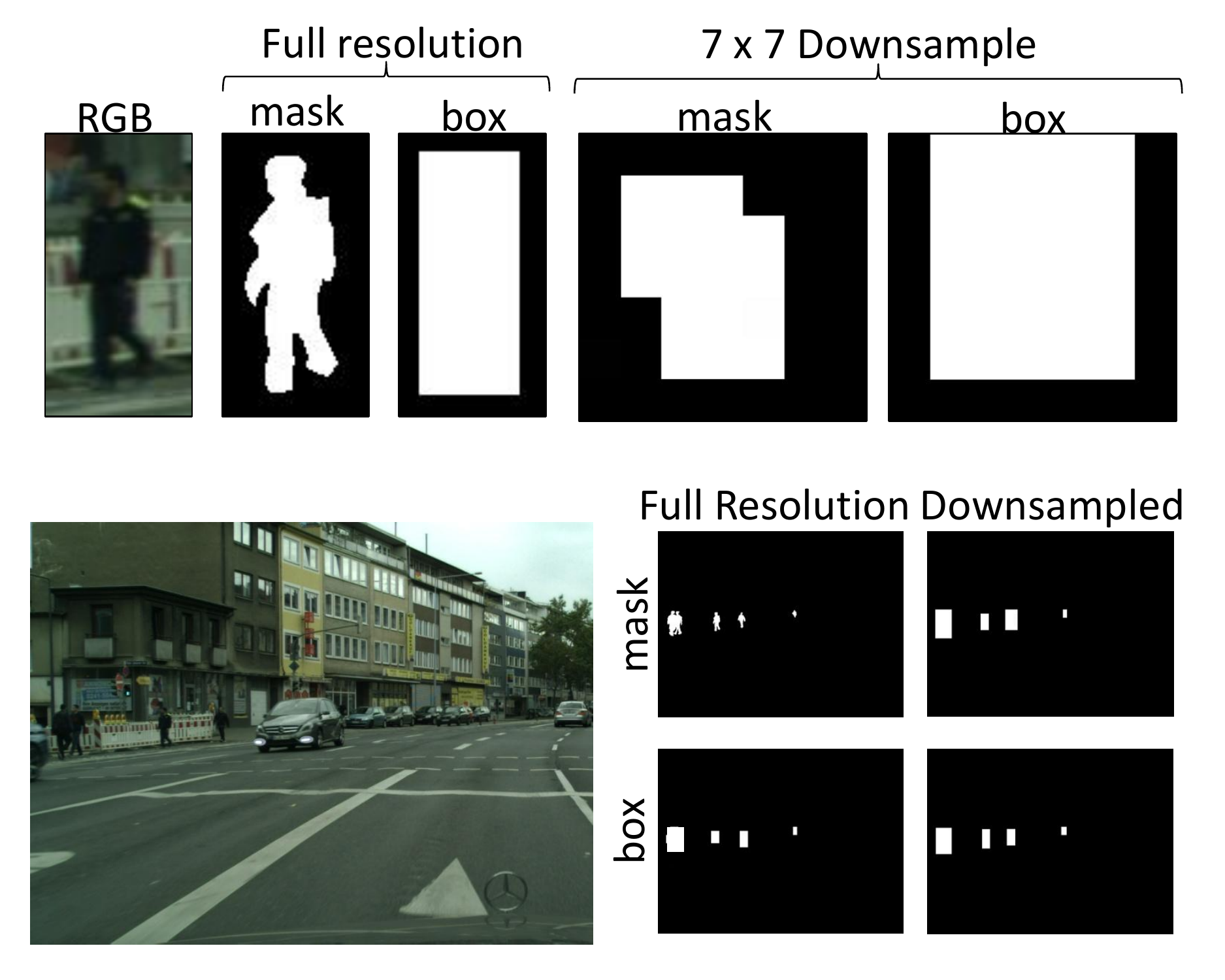}
\end{center}\eqnvspace
   \caption{Visualization of the similarity between pixel-wise segmentation masks (from Cityscapes~\cite{cordts2016cityscapes}) and weak box-based masks when downsampled in both the BCN (top) and RPN (bottom). }
\label{fig:downsample}\figvspace
\end{figure}

We choose to levereage the abundance of bounding box annotations available in popular pedestrian datasets (e.g., Caltech~\cite{dollar2009pedestrian}, KITTI~\cite{Geiger2012CVPR}) by forming weak segmentation ground truth masks. 
Each mask $\mathbf{S}\in\mathbb{R^{W \times H}}$ is generated by labeling all pedestrian box regions as $\mathbf{S}_i=1$, and otherwise background $\mathbf{S}_i=0$. 
In most cases, box-based annotations would be considered too noisy for semantic segmentation. 
However, 
since we place the infusion layer at conv5, which has been pooled significantly,
the differences between box-based annotations and pixel-wise annotations diminish rapidly w.r.t.~the pedestrian height (Fig.~\ref{fig:downsample}).
For example, in the Caltech dataset $68\%$ of pedestrians are less than $80$ pixels tall, which corresponds to $3 \times 5$ pixels at conv5 of the RPN.
Further, each of the BCN proposals are pooled to $7\times7$ at conv5.
Hence, pixel-wise annotations may not offer a significant advantage over boxes at the high levels of pooling our networks undertake.

\Paragraph{Benefits Over Detection:}
A significant advantage of segmentation supervision over detection is its simplicity. 
For detection, sensitive hyperparamters must be set, such as anchor selection and IoU thresholds used for labeling and NMS.
If the chosen anchor scales are too sparse or the IoU threshold is too high, certain ground truths that fall near the midpoint of two anchors could be missed or receive low supervision. 
In contrast, semantic segmentation treats all ground truths indiscriminate of how well the pedestrian's shape or occlusion-level matches the chosen set of anchors. 
In theory, the incorporation of semantic segmentation infusion may help reduce the \textit{sensitivity} of conv1-5 to such hyperparamters.
Furthermore, the segmentation supervision is especially beneficial for the second stage BCN, which on its own would only know \textit{if} a pedestrian is present. 
The infusion of semantic segmentation features inform the BCN \textit{where} the pedestrian is, which is critical for differentiating poorly vs. well-localized proposals. 

%% file: sec_4.tex
\section{Experiments} \label{experiments}

We evaluate our proposed SDS-RCNN on popular datasets including Caltech~\cite{dollar2009pedestrian} and KITTI~\cite{Geiger2012CVPR}. 
We perform comprehensive analysis and ablation experiments using the Caltech dataset. 
We refer to our collective method as SDS-RCNN and our region proposal network as SDS-RPN. 
We show the performance curves compared to the state-of-the-art pedestrian detectors on Caltech in Fig.~\ref{fig:caltech_ROC}.
We further report a comprehensive overview across datasets in Table~\ref{kitti_table}.

\subsection{Benchmark Comparison}

\begin{figure}[t]
\begin{center}
   \includegraphics[width=1\linewidth]{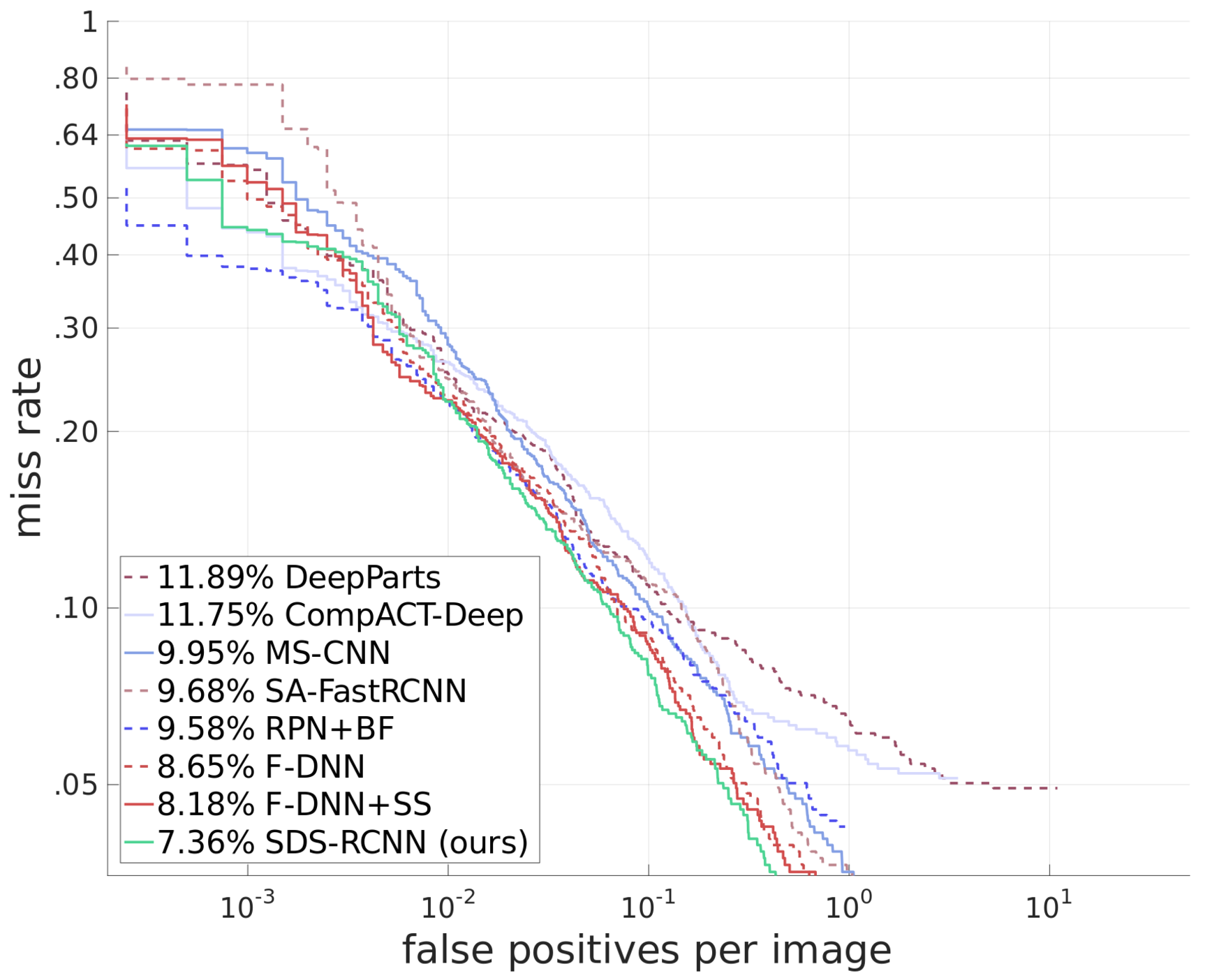}
\end{center}
   \caption{Comparison of SDS-RCNN with the state-of-the-art methods on the Caltech dataset using the \textit{reasonable} setting. }
\label{fig:caltech_ROC}\figvspace
\end{figure}

\Paragraph{Caltech:}
The Caltech dataset~\cite{dollar2009pedestrian} contains $\Sim350$K pedestrian bounding box annotations across $10$ hours of urban driving. 
The log average miss rate sampled against a false positive per image (FPPI) range of $[10^{-2},10^0]$ is used for measuring performance.
A minimum IoU threshold of $0.5$ is required for a detected box to match with a ground truth box.
For training, we sample from the standard training set according to Caltech$10\times$\cite{zhang2015filtered}, which contains $42{,}782$ training images. 
We evaluate on the standard $4{,}024$ images in the Caltech~$1\times$ test set using the \textit{reasonable}~\cite{dollar2012pedestrian} setting, which only considers pedestrians with at least $50$ pixels in height and with less than $35\%$ occlusion.

SDS-RCNN achieves an impressive \textbf{$7.36\%$} miss rate. 
The performance gain is a relative improvement of $23\%$ compared to the best published method RPN+BF ($9.58\%$).  
In Fig.~\ref{fig:caltech_ROC}, we show the ROC plot of miss rate against FPPI for the current top performing methods reported on Caltech. 

We further report our performance using just SDS-RPN (without cost-sensitive weighting, Sec.~\ref{sec:aba}) on Caltech as shown in Table~\ref{kitti_table}.
The RPN performs quite well by itself, reaching $9.63\%$ miss rate while processing images at roughly $3\times$ the speed of competitive methods. 
Our RPN is already on par with other top detectors, which themselves contain a RPN. 
Moreover, the network significantly outperforms other standalone RPNs such as in~\cite{zhang2016faster} ($14.9\%$).
Hence, the RPN can be leveraged by other researchers to build better detectors in the future.

\Paragraph{KITTI:}
The KITTI dataset~\cite{Geiger2012CVPR} contains $\Sim80$K annotations of cars, pedestrians, and cyclists. 
Since our focus is on pedestrian detection, we continue to use only the pedestrian class for training and evaluation.
The mean Average Precision (mAP)~\cite{pascal-voc-2011} sampled across a recall range of $[0,1]$ is used to measure performance.
We use the standard training set of $7{,}481$ images and evaluate on the designated test set of $7{,}518$ images.
Our method reaches a score of $63.05$ mAP on the moderate setting for the pedestrian class. 
Surprisingly, we observe that many models which perform well on Caltech do not generalize well to KITTI, as detailed in Table~\ref{kitti_table}. 
We expect this is due to both sensitivity to hyperparameters and the smaller training set of KITTI ($\Sim6\times$ smaller than Caltech$10\times$). 
MS-CNN~\cite{cai2016unified} is the current top performing method for pedestrian detection on KITTI. 
Aside from the novelty as a multi-scale object detector, MS-CNN augments the KITTI dataset by random cropping and scaling. 
Thus, incorporating data augmentation could alleviate the smaller training set and lead to better generalization across datasets. 
Furthermore, as described in the ablation study of Sec.~\ref{sec:aba}, our weak segmentation supervision primarily improves the detection of unusual shapes and poses (e.g., cyclists, people sitting, bent over). 
However, in the KITTI evaluation, the person sitting class is ignored and cyclists are counted as false positives, hence such advantages are less helpful.

\begin{table}
\begin{center}
\begin{tabular}{|l|c|c|c|c|c|}
\hline
Method & Caltech & KITTI & Runtime \\
\hline\hline
DeepParts~\cite{tian2015deep} & $11.89$ & $58.67$& $1$s \\
CompACT-Deep~\cite{cai2015learning} & $11.75$ & $58.74$ & $1$s\\
MS-CNN~\cite{cai2016unified} & $9.95$ & $73.70$ & $0.4$s \\
SA-FastRCNN~\cite{li2015scale} & $9.68$ & $65.01$ & $0.59$s  \\
RPN+BF~\cite{zhang2016faster} & $9.58$ & $61.29$ & $0.60$s \\
F-DNN~\cite{du2016fused} & $8.65$ & - & $0.30$s  \\
F-DNN+SS~\cite{du2016fused} & $8.18$ & -  & $2.48$s \\
\hline\hline
SDS-RPN (ours) & $9.63$ & -  & $0.13$s \\
SDS-RCNN (ours) & $7.36$ & $63.05$ & $0.21$s \\
\hline
\end{tabular}
\end{center}
\caption{Comprehensive comparison of SDS-RCNN with other state-of-the-art methods showing the Caltech miss rate, KITTI mAP score, and runtime performance.
}
\figvspace
\label{kitti_table}
\end{table}

\Paragraph{Efficiency:}
The runtime performance of SDS-RCNN takes $\Sim0.21$s/image. 
We use images of size $720\times960$ pixels and a single Titan X GPU for computation.
The efficiency of SDS-RCNN surpasses the current state-of-the-art methods for pedestrian detection, often by a factor of $2\times$. 
Compared to F-DNN+SS~\cite{du2016fused}, which also utilizes segmentation cues, our method executes $\Sim10\times$ faster. 
The next fastest runtime is F-DNN, which takes $0.30$s/image with the caveat of requiring multiple GPUs to process networks in parallel.
Further, our SDS-RPN method achieves very competitive accuracy while only taking $0.13$s/image (frequently $\Sim3\times$ faster than competitive methods using a single GPU).

\subsection{Ablation Study}
\label{sec:aba}

\begin{figure*}[t]
\begin{center}
   \includegraphics[width=1\linewidth]{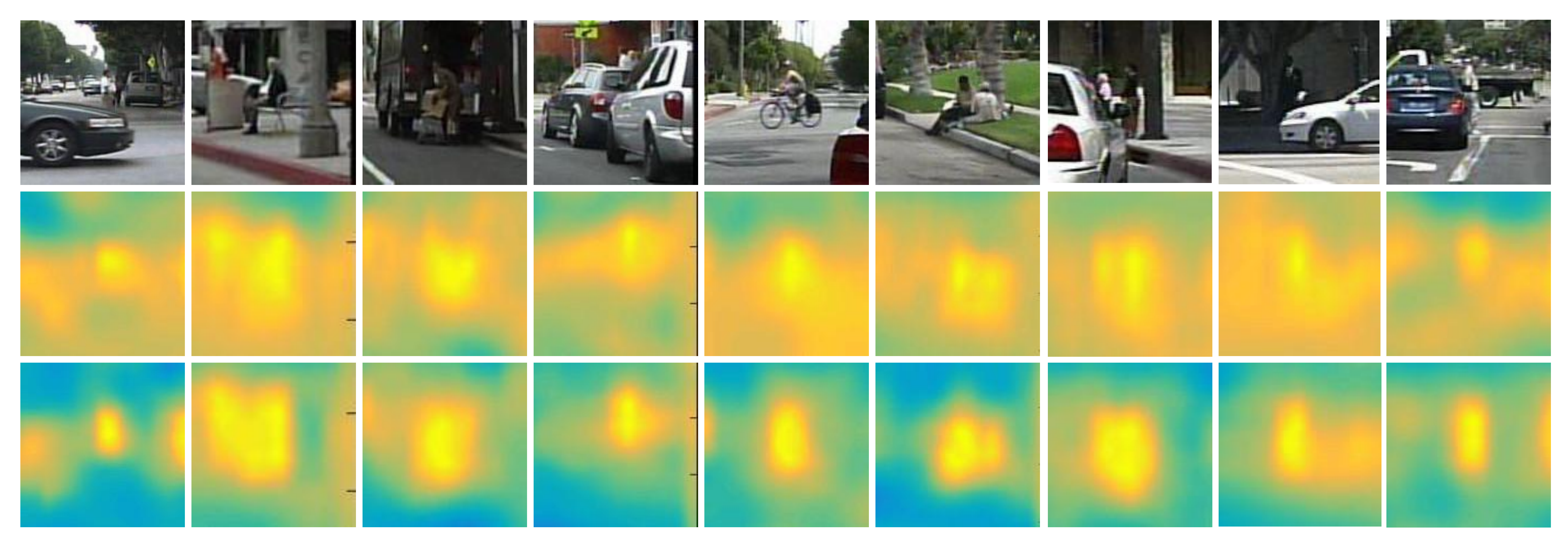}
\end{center}
   \caption{Example error sources which are corrected by infusing semantic segmentation into shared layers. Row $1$ shows the test images from Caltech$1\times$. Row $2$ shows a visualization of the RPN proposal layer using the baseline network which fails on these examples. Row $3$ shows a visualization of the proposal layer from SDS-RCNN, which corrects the errors. Collectively, occlusion and \textit{unusual} poses of pedestrians (sitting, cyclist, bent over) make up for $75\%$ of the corrections, suggesting that the the segmentation supervision naturally informs the shared features on robust pedestrian parts and shape information. 
}
\label{fig:weak_analysis}\figvspace
\end{figure*}

In this section, we evaluate how each significant component of our network contributes to performance using the reasonable set of Caltech~\cite{dollar2009pedestrian}.  
First, we examine the impact of four components: weak segmentation supervision, proposal padding, cost-sensitive weighting, and stricter supervision. 
For each experiment, we start with SDS-RCNN and disable one component at a time as summarized in Table~\ref{ablation_table}.
For simplicity, we disable components globally when applicable.  
Then we provide detailed discussion on the benefits of stage-wise fusion and comprehensively report the RPN, BCN, and fused performances for all experiments.
Finally, since our BCN is designed to not share features with the RPN, we closely examine how sharing weights between stages impacts network diversification and efficiency.

\begin{table}
\begin{center}
\begin{tabular}{|l|c|c|c|}
\hline
Component Disabled & RPN & BCN & Fusion\\
\hline
proposal padding & $10.67$ & $13.09$  & $7.69$\\
cost-sensitive & $9.63$ &  $14.87$  & $7.89$\\
strict supervision & $10.67$ & $17.41$  & $8.71$ \\
weak segmentation & $13.84$ & $18.76$  & $10.41$ \\
\hline
\hline
SDS-RCNN & $10.67$ & $10.98$  & $7.36$ \\
\hline
\end{tabular}
\end{center}
\caption{Ablation experiments evaluated using the Caltech test set. Each ablation experiment reports the miss rate for the RPN, BCN, and fused score with one component disabled at a time.}\figvspace
\label{ablation_table}
\end{table}

\Paragraph{Weak Segmentation:}
The infusion of semantic features into shared layers is the most critical component of SDS-RCNN. 
The fused miss rate degrades by a full $3.05\%$ when the segmentation supervision is disabled, while both individual stages degrade similarly. To better understand the types of improvements gained by weak segmentation, we perform a failure analysis between SDS-RCNN and the ``baseline" (non-weak segmentation) network.
For analysis, we examine the $43$ pedestrian cases which are missed when weak segmentation is disabled, but corrected otherwise. Example error corrections are shown in Fig.~\ref{fig:weak_analysis}. 
We find that $\Sim48\%$ of corrected pedestrians are at least partially occluded.
Further, we find that $\Sim28\%$ are pedestrians in \textit{unusual} poses (e.g., sitting, cycling, or bent over). 
Hence, the feature maps infused with semantic features become more robust to atypical pedestrian shapes.
These benefits are likely gained by semantic segmentation having indiscriminant coverage of all pedestrians, unlike object detection which requires specific alignment between pedestrians and anchor shapes.
A similar advantage could be gained for object detection by expanding the coverage of anchors, but at the cost of computational complexity.


\Paragraph{Proposal Padding:} 
While padding proposals is an intuitive design choice to provide background context (Fig.~\ref{fig:padding}), the benefit in practice is minor.
Specifically, when proposal padding is disabled, the fused performance only worsens from $7.36\%$ to $7.69\%$ miss rate.  
Interestingly, proposal padding remains critical for the individual BCN performance, which degrades heavily from $10.98\%$ to $13.09\%$ without padding.
The low sensitivty of the fused score to padding suggests that the RPN is already capable of localizing and differentiating between partial and full-pedestrians, thus improving the BCN in this respect is less significant. 

\Paragraph{Cost-sensitive:} 
The cost-sensitive weighting scheme used to regularize the importance of large pedestrians over small pedestrians has an interesting effect on SDS-RCNN. 
When the cost-sensitive weighting is disabled, the RPN performance actually improves to an impressive $9.63\%$ miss rate.
In contrast, without cost-sensitive weighting the BCN degrades heavily, while the fused score degrades mildly. 
A logical explanation is that imposing a precedence on a single scale is counter-intuitive to the RPN achieving high recall across \textit{all} scales.
Further, the RPN has the freedom to learn scale-dependent features, unlike the BCN which warps to a fixed size for every proposal.
Hence, the BCN can gain significant boost when encouraged to focus on large pedestrian features, which may be more scale-independent than features of small pedestrians. 

\Paragraph{Strict Supervision:}
Using a stricter labeling policy while training the BCN has a substantial impact on the performance of both the BCN and fused scores. 
Recall that the strict labeling policy requires a box to have IoU $> 0.7$ to be considered foreground, while the standard policy requires IoU $> 0.5$.
When the stricter labeling policy is reduced to the standard policy, the fused performance degrades by $1.35\%$. 
Further, the individual BCN degrades by $6.43\%$, which is on par with the degradation observed when weak segmentation is disabled. 
We examine the failure cases of the strict versus non-strict BCN and observe that the false positives caused by double detections reduce by $\Sim22\%$. 
Hence, the stricter policy enables more aggressive suppression of poorly localized boxes and therefore reduces double detections produced as localization errors of the RPN.

\begin{figure}[t]
\begin{center}
   \includegraphics[width=1\linewidth]{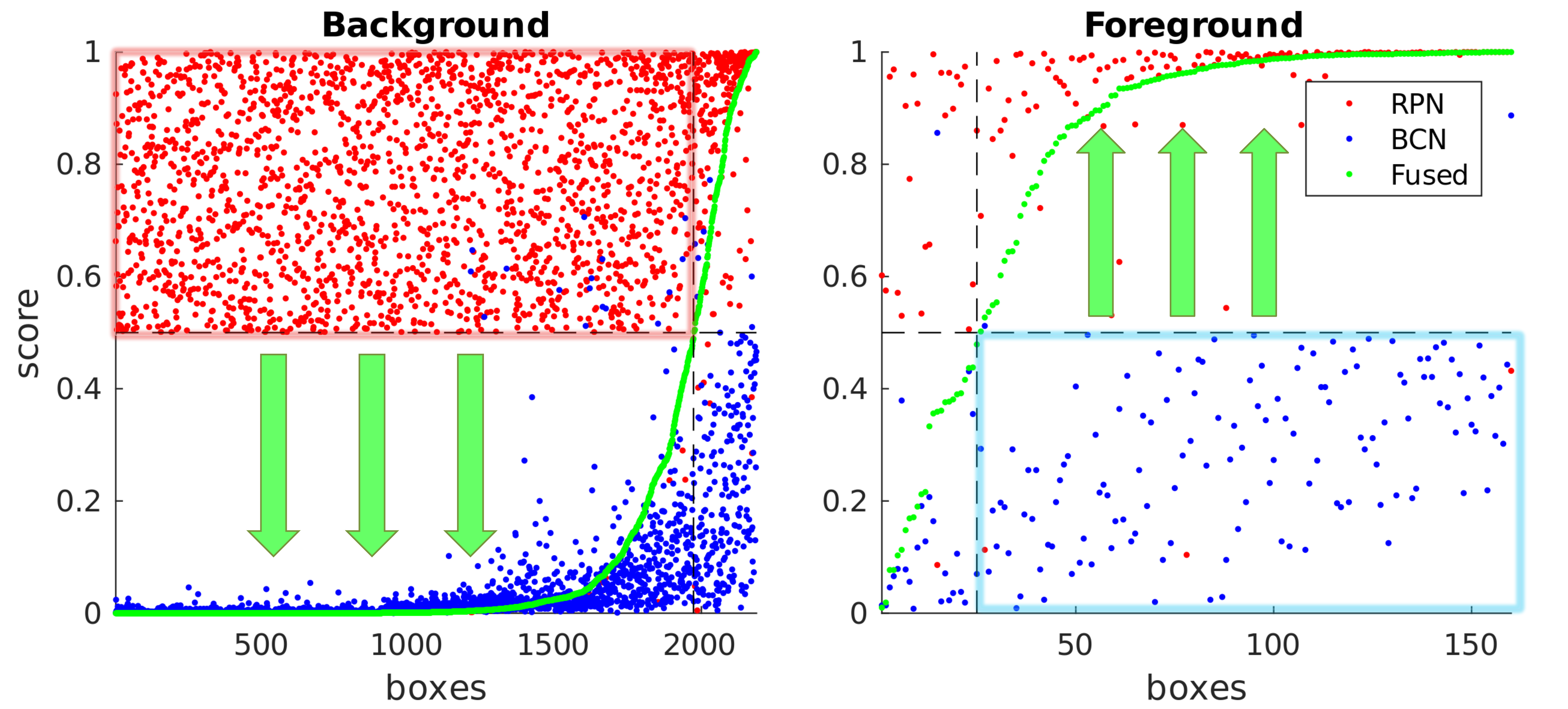}
\end{center}
   \caption{Visualization of the diversification between the RPN and BCN classification characteristics. We plot only boxes which the RPN and BCN of SDS-RCNN disagree on using a threshold of $0.5$. The BCN drastically reduces false positives of the RPN, while the RPN corrects many missed detections by the BCN. }
\label{fig:fusion_plot}\figvspace
\end{figure}

\Paragraph{Stage Fusion:}
The power of stage-wise fusion relies on the assumption that the each network will diversify in their classification characteristics.
Our design explicitly encourages this diversification by using separate labeling policies and training distributions for the RPN and BCN.
Table~\ref{ablation_table} shows that although fusion is useful in every case, it is difficult to anticipate how well any two stages will perform when fused without examining their specific strengths and weaknesses. 

To better understand this effect, we visualize how fusion behaves when the RPN and BCN disagree (Fig.~\ref{fig:fusion_plot}).
We consider only boxes for which the RPN and BCN disagree using a decision threshold of $0.5$.
We notice that both networks agree on the majority of boxes ($\Sim80$K), but observe an interesting trend when they disagree. 
The visualization clearly shows that the RPN tends to predict a significant amount of background proposals with high scores, which are corrected after being fused with the BCN scores.
The inverse is true for disagreements among the foreground, where fusion is able to correct the majority of pedestrians boxes given low scores by the BCN. 
It is clear that whenever the two networks disagree, the fused result tends toward the true score for more than $\Sim80\%$ of the conflicts.


\begin{table}
\begin{center}
\begin{tabular}{|c|c|c|c|c|c|}
\hline
Shared Layer & BCN MR & Fused MR & Runtime \\
\hline\hline
conv5 & $16.24$ & $10.87$ & $0.15$s \\
conv4 & $15.53$ & $10.42$ & $0.16$s \\
conv3 & $14.28$ & $8.66$ & $0.18$s \\
conv2 & $13.71$ & $8.33$ & $0.21$s \\
conv1 & $14.02$ & $8.28$ & $0.25$s \\
\hline\hline
RGB & $10.98$ & $7.36$  & $0.21$s \\
\hline
\end{tabular}
\end{center}
\caption{Stage-wise sharing experiments which demonstrate the trade-off of runtime efficiency and accuracy, using the Caltech dataset. As sharing is increased from RGB (no sharing) to  conv5, both the BCN and Fused miss rate (MR) become less effective. } \figvspace
\label{sharing_table}
\end{table}

\Paragraph{Sharing Features:}
Since we choose to train a separate RPN and BCN, without sharing features, we conduct comprehensive experiments using different levels of stage-wise sharing in order to understand the value of diversification as a trade-off to efficiency.
We adopt the Faster R-CNN feature sharing scheme with five variations differing at the point of sharing (conv1-5) as detailed in Table~\ref{sharing_table}. 
In each experiment, we keep all layers of the BCN except those before and including the shared layer. 
Doing so keeps the effective depth of the BCN unchanged. 
For example, if the shared layer is conv4 then we replace conv1-4 of the BCN with a RoIPooling layer connected to conv4 of the RPN. 
We configure the RoIPooling layer to pool to the resolution of the BCN at the shared layer (e.g., conv4 $\rightarrow14 \times14$, conv5$\rightarrow7\times7$).

We observe that as the amount of sharing is increased, the overall fused performance degrades quickly.
Overall, the results suggest that forcing the networks to share feature maps lowers their freedom to diversify and complement in fusion. 
In other words, the more the networks share the more susceptible they become to redundancies.
Further, sharing features up to conv1 becomes slower than no stage-wise sharing (e.g., RGB).
This is caused by the increased number of channels and higher resolution feature map of conv1 (e.g., $720\times960\times64$), which need to be cropped and warped. 
Compared to sharing feature maps with conv3, using no sharing results in a very minor slow down of $0.03$ seconds while providing a $1.30\%$ improvement to miss rate. 
Hence, our network design favors maximum precision for a reasonable trade-off in efficiency, and obtains speeds generally $2\times$ faster than competitive methods.




%% file: sec_5.tex
\Section{Conclusion}

We present a multi-task infusion framework for joint supervision on pedestrian detection and semantic segmentation. 
The segmentation infusion layer results in more sophisticated shared feature maps which tend to \textit{illuminate} pedestrians and make downstream pedestrian detection easier. 
We analyze how infusing segmentation masks into feature maps helps correct pedestrian detection errors. 
In doing so, we observe that the network becomes more robust to pedestrian poses and occlusion compared to without. 
We further demonstrate the effectiveness of fusing stage-wise scores and encouraging network diversification between stages, such that the second stage classifier can learn a stricter filter to suppress background proposals and become more robust to poorly localized boxes. 
In our SDS-RCNN framework, we report new state-of-the-art performance on the Caltech pedestrian dataset ($23\%$ relative reduction in error), achieve competitive results on the KITTI dataset, and obtain an impressive runtime approximately $2\times$ faster than competitive methods.